\documentclass[wcp]{jmlr}

\usepackage{longtable}

\pdfoutput=1
\usepackage{booktabs}

\usepackage{lineno}
\usepackage{xurl} 

\usepackage{physics}

\makeatletter
\let\Ginclude@graphics\@org@Ginclude@graphics
\makeatother

\title[Improving Denoising Diffusion Models via Simultaneous Estimation of Image and Noise]{Improving Denoising Diffusion Models via Simultaneous Estimation of Image and Noise}

 \author{\Name{Zhenkai Zhang} \Email{zhenkai.zhang@student.unimelb.edu.au}\\
  \Name{Krista A. Ehinger} \Email{kris.ehinger@unimelb.edu.au}\\
  \Name{Tom Drummond} \Email{tom.drummond@unimelb.edu.au}\\
  \addr School of Computing and Information Systems, The University of Melbourne}

\begin{document}

\maketitle

\begin{abstract}
This paper introduces two key contributions aimed at improving the speed and quality of images generated through inverse diffusion processes. The first contribution involves reparameterizing the diffusion process in terms of the angle on a quarter-circular arc between the image and noise, specifically setting the conventional $\displaystyle \sqrt{\bar{\alpha}}=\cos(\eta)$. This reparameterization eliminates two singularities and allows for the expression of diffusion evolution as a well-behaved ordinary differential equation (ODE). In turn, this allows higher order ODE solvers such as Runge-Kutta methods to be used effectively. The second contribution is to directly estimate both the image ($\mathbf{x}_0$) and noise ($\mathbf{\epsilon}$) using our network, which enables more stable calculations of the update step in the inverse diffusion steps, as accurate estimation of both the image and noise are crucial at different stages of the process. Together with these changes, our model achieves faster generation, with the ability to converge on high-quality images more quickly, and higher quality of the generated images, as measured by metrics such as Fréchet Inception Distance (FID), spatial Fréchet Inception Distance (sFID), precision, and recall. \textit{Code is available at} \url{https://github.com/Fredy-Zhang/arcDiff}.
\end{abstract}


\section{Introduction}
Diffusion models~\cite{sohl2015deep}; \cite{ho2020denoising}, which have emerged as powerful tools, are capable of generating synthetic images that exhibit both high quality and diversity. These models leverage the concept of gradual image refinement through a diffusion process, where noise is sequentially transformed into realistic images. The ability to generate visually appealing and realistic images has found applications in various fields, such as data augmentation, computer vision, and image synthesis.

Significant progress has been made in recent years in advancing diffusion models, either by improving the empirical performance~\cite{nichol2021improved}; \cite{song2020denoising}; \cite{song2020improved} or by extending the capabilities of the models from a theoretical perspective~\cite{lu2022maximum}; \cite{lu2022dpm}; \cite{song2021maximum}; \cite{song2020score}, leading to improvements in image quality and generation capabilities. However, the current diffusion models face a limitation in terms of time inefficiency during the inference process, particularly in models that utilize noise as the target object~\cite{ho2020denoising}; \cite{nichol2021improved}; \cite{song2020denoising}. This inefficiency is due to the substantial amount of time required in the initial stages of sampling, where the model progresses from pure noise to low-quality images. Meanwhile, models that use images as the target object~\cite{bansal2022cold}, do not encounter the same issues as noise-based diffusion models, as they can achieve a faster transition from pure noise to low-quality images. However, these models still face the challenge of directly estimating the image, which becomes increasingly difficult at the final stage of the diffusion process when the input is dominated by noise. Consequently, this difficulty leads to a lower performance of the final result than the noise-based diffusion model.

\begin{figure}[h]
\begin{center}
\includegraphics[width=1.0\textwidth]{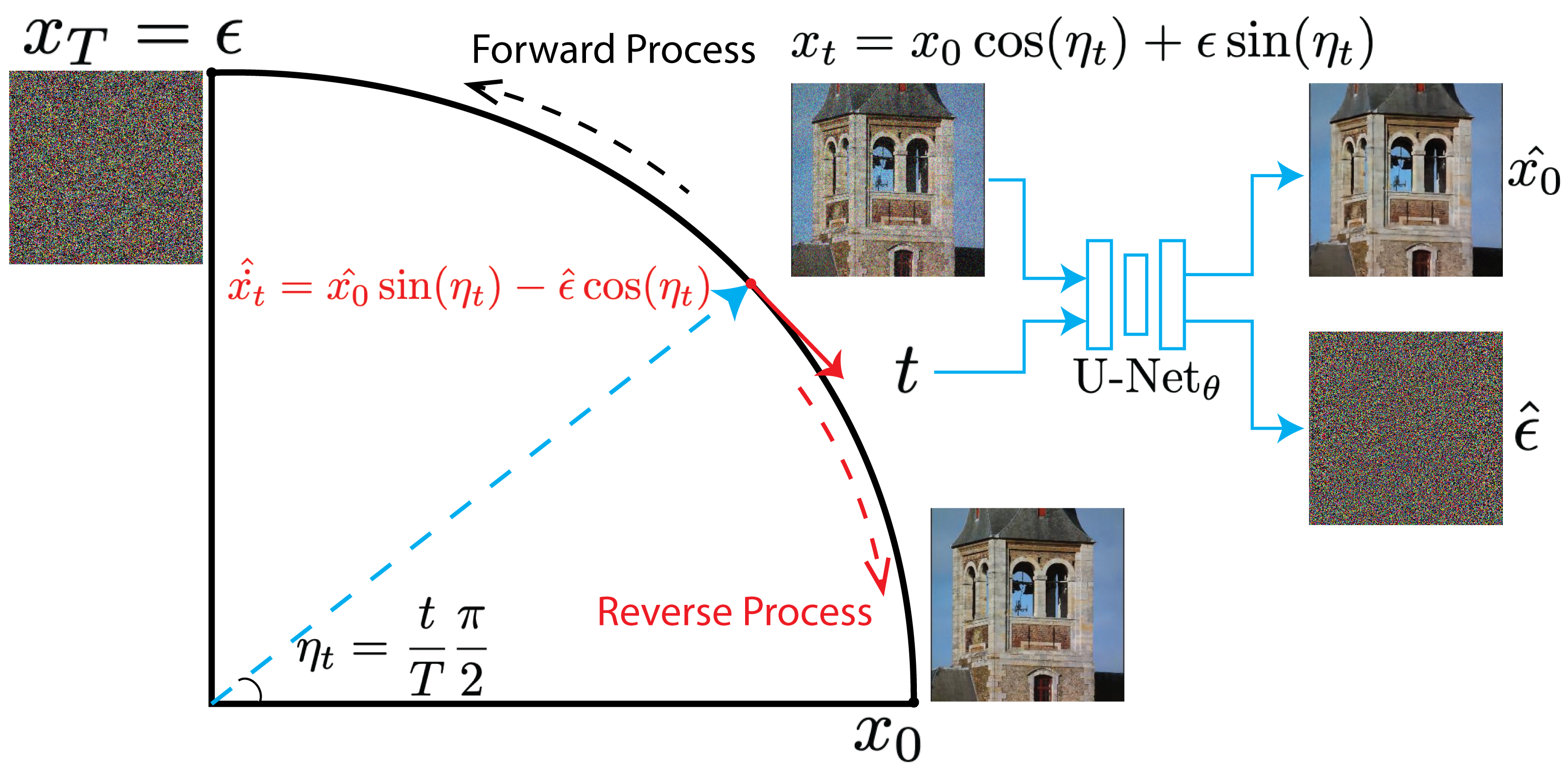}
\caption{The overview of our model architecture and concept. It depicts how our system diffuses from input images to images with noise, eventually leading to pure noise. The figure also showcases our sampling process, which utilizes gradients to generate images from noise.}\label{fig:overview}
\end{center}
\end{figure}

To address these limitations, in this paper, we propose a novel approach that combines the advantages of both noise-targeted training and image-targeted training in the diffusion process. By simultaneously incorporating noise and actual images as training objectives, our model aims to overcome the aforementioned limitations and achieve superior performance in terms of both result quality and diversity. \figureref{fig:overview} depicts our model architecture and concepts.

Overall, our framework proposes several practical and theoretical contributions,

\begin{itemize}
    \item Our approach develops a new and efficient parameterization technique that results in an innovative noise scheduler. In \sectionref{exp:al} of the ablation experiments, we describe that this approach improves the overall quality of generated images while reducing the number of steps needed to produce high-quality results.

    \item Our proposed model significantly improves over traditional models that only estimate images or noise separately. Our model is trained to simultaneously predict both noise and images, which improves the estimate of the gradient at each timestep. We use gradients as a loss function to update the model, which enhances controllability during image generation. Additionally, we utilize gradient information to optimize the generation process during the reversed process. Our model obtains higher quality and more realistic results by calculating the gradient information of the current state and using that information to update the steps more accurately and stable, and the results are shown in \sectionref{exp:re}.

\end{itemize}

\section{Background and Related Work}
In this section, a brief introduction is provided regarding the relevant work that our model is built upon.

The concept of diffusion models \cite{sohl2015deep};\cite{ho2020denoising} is rooted in non-equilibrium thermodynamics, wherein the diffusion and reverse processes act as Markov chains. Unlike other approaches to generative modelling such as Variational Autoencoders (VAEs, \cite{kingma2013auto}), normalizing flows (\cite{rezende2015variational}), or Generative Adversarial Networks (GANs, \cite{goodfellow2020generative}), the purpose of the diffusion model is to describe the distribution of the data by gradually reducing the noise or uncertainty of the input, and the latent space has the same dimensionality as the input space. 

\subsection{Noise-based diffusion models}
The majority of diffusion models \cite{Kong2020DiffWaveAV}; \cite{song2020denoising}; \cite{Nichol2021ImprovedDD}; \cite{SanRoman2021NoiseEF}; \cite{Rombach2021HighResolutionIS} are trained by adding Gaussian noise to the images to generate noisy samples, and feeding them into the U-Net network~\cite{ronneberger2015u} for predicting the added noise. This process is effective because incorporating noise prediction into the training procedure can effectively disentangle the noise from the underlying signal, allowing for a more accurate and robust generation of samples, and alleviating the burden of explicitly modelling complex data patterns. The first widely followed and used model is the Denoising diffusion probabilistic models (DDPMs, \cite{ho2020denoising}), latent variable models that utilise a Markov chain concept. These models have learned Gaussian transitions that begin with a Normal distribution $p(x_T) \sim \mathcal{N}(x_T; \boldsymbol{0}, \boldsymbol{I})$ in the following form:
\begin{equation}
    \displaystyle  p_\theta(x_0) := \int p_\theta(x_{0:T})dx_{1:T}, \;\;\; \displaystyle p_\theta(x_{0:T}) := p(x_T) \prod_{t=1}^{T} p_\theta(x_{t-1}|x_t)
\end{equation}
where $x_1,x_2,...,x_T$ keep the same shape with $x_0$, $\theta$ means the weights of neural networks, commonly using the U-net structure, and the $p_\theta(x_{0:T})$ is defined as the reverse process. The parameters $\theta$ are trained to match the data distribution $q(x_0)$ by maximizing a variational lower bound by optimizing the negative log-likelihood.
\begin{equation}
    \displaystyle L_{VLB} = \max_{\theta} \mathbb{E}_{q(x_{0:T})} \displaystyle \left[ \log \frac{q(x_{1:T}|x_0)}{p_\theta(x_{0:T})} \right] \ge \max_\theta \left[ -\mathbb{E}_{q(x_0)}\left[ \log p_\theta(x_0) \right] \right]
\label{eq:2}
\end{equation}
where the $q(x_{1:T}|x_0)$ follows a Markov chain that systematically introduces Gaussian noise to the data, following a pre-determined variance schedule ${\beta_t \in (0,1)}^T_{t=1}$, named as the forward process or diffusion process. \cite{ho2020denoising} defines $\beta_t$ as a sequence of linearly increasing constants from $\beta_1=0.0001$ to $\beta_T=0.02$.
\begin{equation}
    q(x_{1:T}|x_0) := \prod_{t=1}^{T} q(x_t|x_{t-1}), \; \text{where}\; q(x_t|x_{t-1}) \sim \mathcal{N}(x_t;\sqrt{1-\beta_t}x_{t-1}, \beta_t \boldsymbol{I})
\label{eq:3}
\end{equation}
Then, by employing the reparameterization trick~\cite{rezende2015variational}, the conditional distribution $q(x_t|x_{t-1})$ can be expressed as a closed form.
$$\displaystyle x_t = \sqrt{1-\beta_t} \mathbf{x_{t-1}} + \sqrt{\beta_t} \mathbf{\epsilon_{t-1}},\text{ where }\mathbf{\epsilon_{t-1}} \sim \mathcal{N}(\mathbf{0},\mathbf{I})$$ 
Let $\alpha_t = 1-\beta_t$ and $\bar{\alpha}_t=\prod_{i=1}^{t} \alpha_i$,
\begin{equation}
    x_t = \sqrt{\bar{\alpha}_t} \mathbf{x_0} + \sqrt{1-\bar{\alpha}_t} \mathbf{\epsilon}, \; \mathbf{\epsilon} \sim \mathcal{N}(\mathbf{0}, \mathbf{I}),
\label{eq:4}
\end{equation}
Among them, the linear design of the $\beta_t$ in \cite{ho2020denoising} causes most of the values in $\bar{\alpha}_t$ to be concentrated near 0, as shown in the \figureref{fig:3}, thus causing inefficiencies in the sampling process. 

Combining \equationref{eq:4}, we can simplify the objective function defined by \equationref{eq:2} to,
\begin{equation}
    \displaystyle \mathbb{E}_{\mathbf{x_0,\epsilon}} \left[ C \| \mathbf{\epsilon}_\theta(\sqrt{\bar{\alpha}_t} \mathbf{x_0} + \sqrt{1-\bar{\alpha}_t} \mathbf{\epsilon}, t) - \mathbf{\epsilon} \| \right], \text{ where C is an constant.}
\label{eq:5}
\end{equation}
However, DDPM~\cite{ho2020denoising} suffers from limitations in inference due to its Markov chain-based framework, leading to computational burden and extended inference times. 

By addressing the limitations of DDPM, Denoising Diffusion Implicit Models (DDIM,~\cite{song2020denoising}) improve inference by utilizing non-Markovian methods and implicit models to reduce the computational burden and enable faster and more efficient inference, without changing the training process. \cite{song2020denoising} proposes that comparing the FID score of DDPM, the DDIM model can obtain lower fid values when the number of steps is below 1000, but the DDPM model achieves better results when the number of steps is equal to 1000.

The forward process of \equationref{eq:3} from DDPM is that the $x_t$ only rely on $x_{t-1}$ based on Markovian, and the DDIM derives such forward process from Bayes' rule,
\begin{equation*}
    q_\sigma(x_t|x_{t-1},x_0) = \frac{q_\sigma(x_{t-1}|x_t, x_0)q_\sigma(x_t|x_0)}{q_\sigma(x_{t-1}|x_0)}
\end{equation*}
where the hyper-parameter $\sigma$ is utilized to regulate the stochastic nature of the forward process, then $q_{\sigma}(\mathbf{x_{t-1}}|\mathbf{x_t},\mathbf{x_0})$ defines as $\displaystyle \mathcal{N}(\sqrt{\bar{\alpha}_{t-1}}\mathbf{x_0} + \sqrt{1-\bar{\alpha}_{t-1} - \sigma_t^2}\frac{\mathbf{x_t}-\sqrt{\bar{\alpha}_t} \mathbf{x_0}}{\sqrt{1-\bar{\alpha}_t}}, \sigma_t^2\mathbf{I})$. For the $\sigma_t=0$ setting in DDIM, the forward process is deterministic except for $t = 1$, given the values of $\mathbf{x}_{t-1}$ and $\mathbf{x}_0$.

Then, the reverse process from sample $\mathbf{x}_t$ to the generated sample $\mathbf{x}_{t-1}$ can be defined as (setting the $\sigma=0$)
\begin{equation}
    \mathbf{x}_{t-1} = \sqrt{\displaystyle \bar{\alpha}_{t-1}} \hat{\mathbf{x}_0} + \sqrt{1-\displaystyle \bar{\alpha}_{t-1}}\epsilon_\theta(\mathbf{x}_t), \text{ where } \hat{\mathbf{x}_0}=\frac{\mathbf{x}_t-\sqrt{1-\bar{\alpha}_t}\mathbf{\epsilon}_\theta(\mathbf{x}_t)}{\sqrt{\bar{\alpha}_t}}
\label{eq:6}
\end{equation}

Meanwhile, there are various methods \cite{Song2020ScoreBasedGM}; \cite{Lu2022DPMSolverAF}; \cite{Lu2022DPMSolverFS} that share a similar motivation with noise-based diffusion models, but they rely on score matching during the reverse process. However, such type of diffusion models are known to have certain limitations, such as difficulty in initial stage learning and limited control over sample generation.

\subsection{Image-based diffusion models}
Some models use a different approach to train generative models. Instead of predicting noise, they train the models to predict images directly \cite{bansal2022cold}; \cite{Hoogeboom2022BlurringDM}. This method simplifies the training process by focusing on learning patterns and structures within input images. By predicting images, these models take advantage of the inherent advantages of images, such as spatial coherence and semantic information, to create high-quality samples. This approach can speed up the learning process, as the initial prediction stages provide meaningful visual information for generating samples. The aim of using image prediction is to capture the essence of the underlying data distribution and generate visually appealing and coherent outputs. 

Cold Diffusion~\cite{bansal2022cold}, as the first model to introduce this concept into diffusion models, defines \equationref{eq:3} as the broader concept of "degradation", which is defined as $\mathbf{x}_0$ by operation $D$ with the timestep $t$, denoted as $\mathbf{x}_t = D(\mathbf{x}_0, t)$. 

Correspondingly, the reverse process is also defined as the "restoration", which is an operation to approximately invert operation $D$, denoted as $R$, implemented by a neural network with $theta$ as the parameter.
\begin{equation*}
    R_\theta(\mathbf{x}_t, t) \simeq \mathbf{x}_0.
\end{equation*}
Therefore this objective function becomes,
\begin{equation}
    \min_\theta \mathbb{E} \left[ \| R_\theta(D(\mathbf{x}_0,t) - \mathbf{x}_0 \| \right]
\label{eq:7}
\end{equation}
The sampling for Cold Diffusion can be defined as
\begin{equation*}
    \mathbf{x}_{t-1} = D(\hat{\mathbf{x}_0}, t-1).
\end{equation*}
When using noise-based degradation $D(\mathbf{x}_0,t)=\sqrt{\bar{\alpha_t}} \mathbf{x}_0+\sqrt{1-\bar{\alpha_t}} \mathbf{\epsilon}$, the sampling process becomes deterministic, which is similar to \equationref{eq:6} defined in DDIM~\cite{song2020denoising}, 
\begin{equation}
    \mathbf{x}_{t-1} = \sqrt{\bar{\alpha}_{t-1}} R_\theta(\mathbf{x}_t,t)+\sqrt{1-\bar{\alpha}_{t-1}}\hat{\mathbf{\epsilon}}(\mathbf{x}_t,t), \;\; \hat{\mathbf{\epsilon}}(\mathbf{x}_t,t) = \frac{\mathbf{x}_t - \sqrt{\bar{\alpha}_t}R_\theta(\mathbf{x}_t,t)}{\sqrt{1-\bar{\alpha}_t}}
\label{eq:8}
\end{equation}

Although image-based diffusion models are useful, they have certain limitations. Specifically, they struggle to capture long-range dependencies and can be computationally complex. Hence, we introduce a novel architecture to eliminate the limitations of noise-based or image-based models while retaining the advantages of both. 

\section{Methodology}
Through examining various categories of diffusion models, it becomes apparent that there are numerous opportunities for enhancing and improving these models. Firstly, proposing a more effective noise scheduler in \sectionref{meth:1}. Secondly, pursuing the simultaneous estimation of both images and noise in \sectionref{meth:2}. Lastly, utilizing gradient descent to improve the stability of the sampling process in \sectionref{meth:3}. 

\subsection{Noise Scheduler}
\label{meth:1}
In the forward process, DDPM, DDIM and noise-based Cold Diffusion all rely on \equationref{eq:4} with the square root parameterization. This definition leads to two singularities at $t=0$ and $t=T$ ($\tfrac{d x_t}{d \bar{\alpha}} \rightarrow \infty$) and also causes more steps to be required in the reverse diffusion process.

We propose changing the parameterization to:
\begin{equation}
    \mathbf{x}_t = \cos(\eta_t) \mathbf{x}_0 + \sin(\eta_t) \mathbf{\epsilon}. \text{ where } \eta_t = \frac{t}{T} \frac{\pi}{2},\; t \in \{0,1,2,...,T\}.
\label{eq:9}
\end{equation}
where mapping the $\sqrt{\bar{\alpha}_t}$ to $\cos(\eta_t)$ and $\sqrt{1-\bar{\alpha}_t}$ to $\sin(\eta_t)$.

The benefit of our new parameterization approach is to avoid the problem of singularities that existed in previous methods. This has the benefit of allowing us to express the reverse diffusion processes as an ordinary differential equation (ODE) in continuous time, as shown in Fig \ref{fig:overview}. In turn, this enables us to replace the commonly used Euler step method with higher order ODE solvers such as Runge-Kutta methods.

Recalling the parameterization equation in DDIM defined in \equationref{eq:4}, there are two singularities at $t=0$ and $t=T$ when computing the gradient for this formula,
\begin{gather*}
    \displaystyle \dv{\mathbf{x}_t}{\bar{\alpha}_t} = \dv{(\sqrt{\bar{\alpha}_t} \mathbf{x_0} + \sqrt{1-\bar{\alpha}_t} \mathbf{\epsilon})}{\bar{\alpha}_t} = \frac{1}{2\sqrt{\bar{\alpha}_t}}\mathbf{x}_0 - \frac{1}{2\sqrt{1-\bar{\alpha}_t}}\mathbf{\epsilon}, \\
    \text{When } t=0, \;\bar{\alpha}_t = 0, \; \left.\dv{\mathbf{x}_t}{\bar{\alpha}_t}\right|_{\bar{\alpha}_t = 0} \to +\infty, \\
    \text{When } t=T, \;\bar{\alpha}_t = 1, \; \left.\dv{\mathbf{x}_t}{\bar{\alpha}_t}\right|_{\bar{\alpha}_t = 1} \to -\infty.
\end{gather*}
However, our new parameterization formula defined in \equationref{eq:9} can remove the singularities,
\begin{gather*}
    \dv{x_t}{\eta_t} = \dv{(\cos(\eta_t) \mathbf{x}_0 + \sin(\eta_t) \mathbf{\epsilon})}{\eta_t} = -\sin(\eta_t)\mathbf{x}_0 + \cos(\eta_t) \mathbf{\epsilon}, \\
    \text{When } t=0, \;\eta_t = 0, \; \left.\dv{\mathbf{x}_t}{\eta_t}\right|_{\eta_t=0} = \mathbf{\epsilon}, \\
    \text{When } t=T, \;\eta_t = \frac{\pi}{2}, \; \left.\dv{\mathbf{x}_t}{\eta_t}\right|_{\eta_t=\frac{\pi}{2}} = -\mathbf{x}_0.
\end{gather*}

With the new parameterization defined in  \equationref{eq:9}, we can back-derive an equivalent schedule for $\bar{\alpha}_t$ and $\beta_t$.
\begin{equation*}
    \bar{\alpha}_t = \cos^2(\eta_t)
\end{equation*}
\begin{equation*}
    \beta_t = 1 - \frac{\cos^2(\eta_t)}{\cos^2(\eta_{t-1})}
\end{equation*}

\begin{figure*}[h]
  \centering
  \begin{minipage}[b]{0.49\textwidth}
    \centering
    \includegraphics[width=0.9\textwidth]{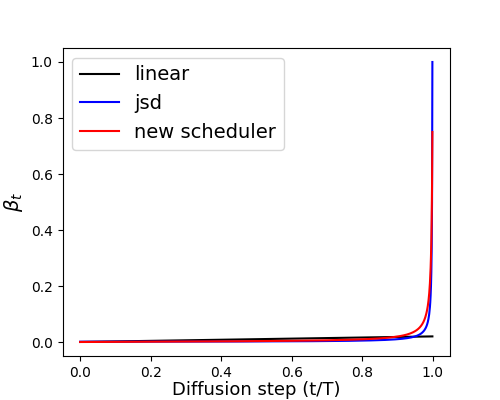}
    \caption{The $\beta_t$ schedule changes with increasing diffusion steps.}
    \label{fig:2}
  \end{minipage}\hfill
  \begin{minipage}[b]{0.49\textwidth}
    \centering
    \includegraphics[width=0.9\textwidth]{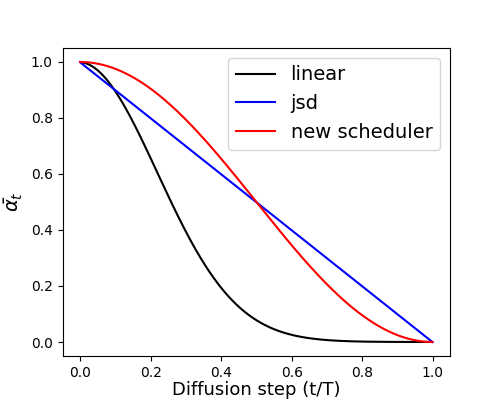}
    \caption{The $\bar{\alpha}_t$ changes with increasing diffusion steps. }
    \label{fig:3}
  \end{minipage}
\end{figure*}

In \figureref{fig:2} and \figureref{fig:3}, comparing the traditional linear noise schedule \cite{ho2020denoising} and JSD schedule, which is defined as $\displaystyle \beta_t = \frac{1}{t},\; t\in \{1,2,3,..., T\}$, we found that our schedule makes the forward process smoother and does not cause the problem of information loss due to sudden drops as in the linear scheme. In practice, we also found that using this schedule would be beneficial in balancing the contribution of image loss and noise loss, in Section \ref{meth:2}.

\subsection{Simultaneous Estimation of Image and Noise}
\label{meth:2}

In the objective function, DDPM~\cite{ho2020denoising} and DDIM~\cite{song2020denoising} train the generative models by estimating the noise, and Cold Diffusion \cite{bansal2022cold} fits the generative model by predicting the image directly. 

\begin{figure*}[h]
\begin{center}
\includegraphics[width=0.9\textwidth]{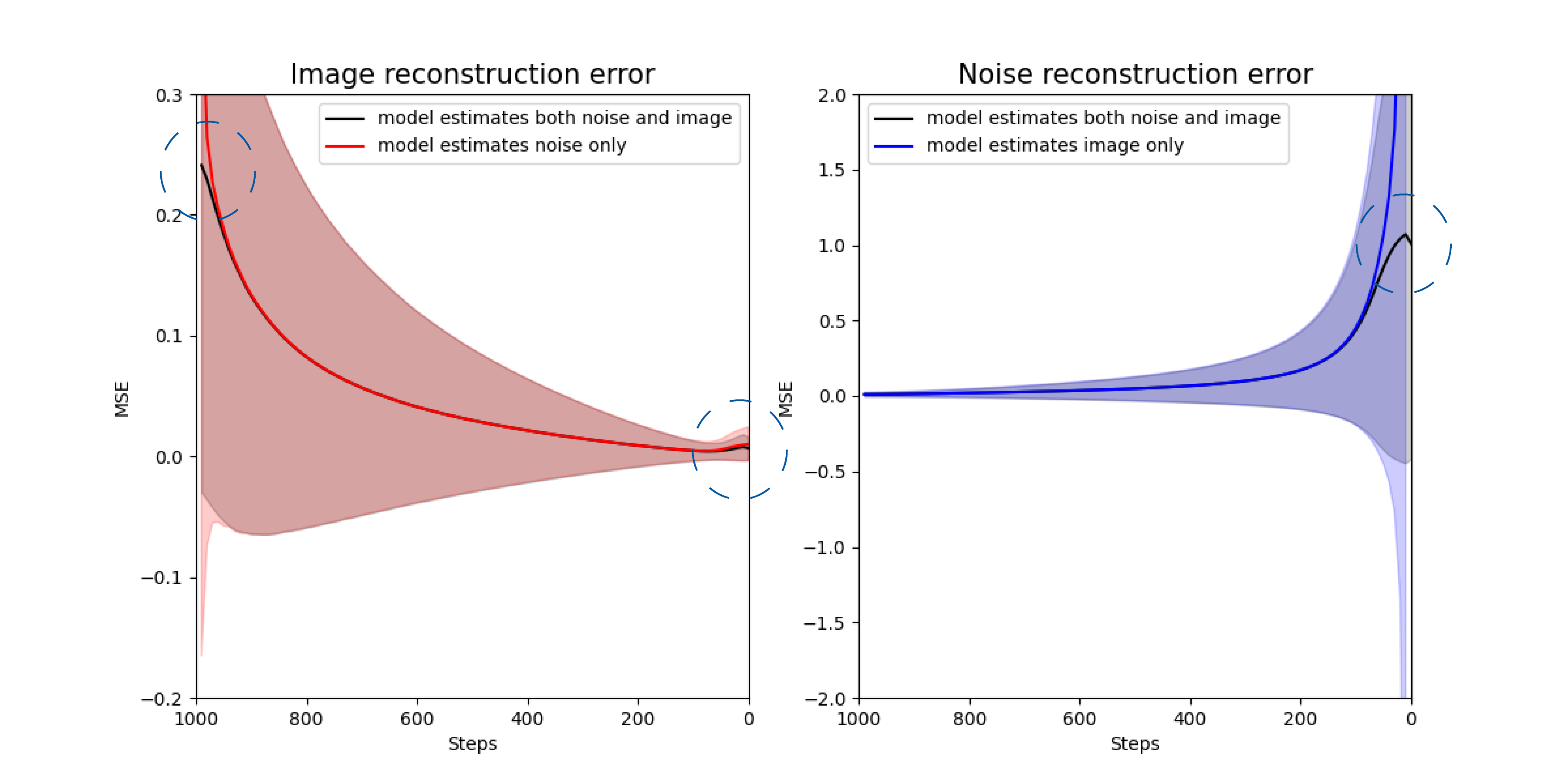}
\caption{Image reconstruction error (left) and noise reconstruction error (right) at inference for a model that estimates both noise and image, compared to models that estimate only one of these. Lines indicate average MSE and the shaded area indicates the standard variance of MSE.} \label{2:fig}
\end{center}
\end{figure*}

Noise-based diffusion models offer certain advantages, particularly in the later stages of sampling where the process becomes simpler and clearer due to the subtraction of noise. However, starting from pure noise during the initial stages of sampling can pose learning challenges because of the presence of noise-dominated data. In contrast, image-based diffusion models exhibit the opposite characteristics. In the initial stages of sampling, they begin with actual images, which facilitates the learning process by providing meaningful information from the start. However, during the later stages of sampling, the complexity increases as the model needs to manipulate and transform the images to generate new samples. Therefore, by simultaneously predicting both noise and images, we aim to overcome the limitations of each approach and achieve a more controllable sampling process. By utilizing joint prediction, we can harness the benefits offered by both noise-based and image-based diffusion models. This approach provides improved control and flexibility throughout the sampling process, as the \figureref{2:fig} shows. 


As shown in \figureref{2:fig}, during the initial steps of the inference stage, the image reconstruction error is lower when estimating both the image and noise, compared to estimating the noise only and using it to derive the image. In the final steps of the inference stage, both image and noise reconstruction errors are lower when both are estimated together, compared to models that estimate image only or noise only.

Hence, by combining \equationref{eq:5} and \equationref{eq:7}, we derive the desired loss function for our model.
\begin{equation*}
    \min_\theta \mathbb{E}\left[ \|\mathbf{R}_\theta(\mathbf{x}_t, t) - \mathbf{x}_0 \| + \|\mathbf{\epsilon}_\theta(\mathbf{x}_t, t) - \mathbf{\epsilon} \| \right]
\end{equation*}

\subsection{Sampling with Gradient Update}
\label{meth:3}
From Figure \ref{fig:overview}, we can conceptualize the entire diffusion process as the progress of starting from an initial image $\mathbf{x}_0$ and progressively reducing the level of the image along a curve until obtaining pure noise $\mathbf{\epsilon}$. Hence, we can consider discovering an image from noise as an iterative optimization process that implements gradient descent to seek the solution. We take the gradient of \equationref{eq:9} to get the ground-truth gradient.
\begin{equation}
    \dot{\mathbf{x}}_t = \eta_t' \left[-\sin(\eta_t)\mathbf{x}_0 + \cos(\eta_t) \epsilon  \right].
\end{equation}
Then, the estimated gradient is,
\begin{equation}
    \displaystyle \widehat{\dot{\mathbf{x}}}_t = \eta_t' \left[-\sin(\eta_t) \mathbf{R}_\theta(\mathbf{x}_t,t) + \cos(\eta_t) \mathbf{\epsilon}_\theta(\mathbf{x}_t,t)  \right]. 
\label{eq:11}
\end{equation}

In practice, we have observed that incorporating the gradient of the loss as part of the objective function significantly improves the performance of the model. Then, we obtain the final objective function,
\begin{equation}
    \min_\theta \mathbb{E}\left[ \|\mathbf{R}_\theta(\mathbf{x}_t, t) - \mathbf{x}_0 \| + \|\mathbf{\epsilon}_\theta(\mathbf{x}_t, t) - \mathbf{\epsilon} \| + \gamma \|\widehat{\dot{\mathbf{x}}} - \dot{\mathbf{x}} \| \right]
\end{equation}
where the $\gamma$ is used to control the weight of gradient loss.

With the gradient update approach, our sampling process is,
\begin{equation}
    \mathbf{x}_{t-1} = \mathbf{x}_t - \Delta_t \widehat{\dot{\mathbf{x}}_t}
\end{equation}
In practice, we can use techniques like Second Order Runge-Kutta (RK2) and Fourth Order Runge-Kutta (RK4) \cite{hairer1993solving}, to extend the gradient function defined in \equationref{eq:11} to improve performance.

\section{Experiments Setup}

\paragraph{Baseline Models.}
In our experiments, we use the noise-based diffusion models DDPM~\cite{ho2020denoising} and DDIM~\cite{song2020denoising} and the image-based diffusion model Cold Diffusion~\cite{bansal2022cold} as baseline models. By comparing these two types of models, we can validate the improvements in performance that our contributions and model bring to diffusion models. However, for the Cold Diffusion model~\cite{bansal2022cold}, we found that this model only allows for full-step inference. In other words, if our model is trained with a total of 200 steps, we can only set it to 200 steps during the inference process to achieve optimal performance. Comparing to \equationref{eq:6} in DDIM~\cite{song2020denoising} and \equationref{eq:8} in Cold Diffusion~\cite{bansal2022cold}, we can conclude that they are from similar ideas, the only difference is that one is to estimate the noise and the other is to predict the image. Models that predict noise tend to perform better than models that predict images when the sampling steps are large. Therefore, we decided not to include it in the experiment for comparison.

\paragraph{Datasets.} We primarily evaluated the performance of our model using three datasets of varying scales, CIFAR-10 ($32\times32$, \cite{krizhevsky2009learning}), CelebA ($178\times218$, \cite{liu2015faceattributes}), and LUSH ($256\times256$, \cite{yu2015lsun}).
For the UNet architecture model, achieving good results often requires input images to have consistent width and height. To maintain consistency with models' data preprocessing in DDIM~\cite{song2020denoising}, we performed cropping on the CelebA dataset, resizing the original images from 178x218 to 64x64 \cite{song2020denoising}. Additionally, in the LUSH dataset, we chose for evaluation a widely used subset of outdoor church images which have dimensions 256x256.

\paragraph{Evaluation Metrics.}  We used four evaluation metrics to gauge the models' performances. The FID~\cite{heusel2017gans} metric is utilized to determine the divergence between the distribution of generated images and the distribution of actual images. Similarly, sFID~\cite{nash2021generating} is also used to measure this divergence while maintaining spatial information of the images. Additionally, precision indicates the fraction of the generated images that match reality, and recall measures the coverage of the training data manifold by the generator~\cite{kynkaanniemi2019improved}.

\paragraph{Implementation Details.} Our proposed method is built upon the DDIM model~\cite{song2020denoising} as the base model, and we use the same trained model with $T=1000$ with the objective function from \equationref{eq:5} for each dataset. To mitigate the potential confusion caused by varying iterations during model training, in our ablation experiments, we employ models with parameters approximating the number of iterations used in the pre-training model. This approach ensures that the compared models have a similar number of iterations, reducing the confounding factor of iteration differences in performance evaluation. In our experiments, we utilized pre-trained models from DDPM for LUSH Church and CIFAR10 datasets~\cite{ho2020denoising}, as well as a pre-trained model from DDIM for the CelebA dataset~\cite{song2020denoising}, as references.


\section{Experiments Results}
\label{exp:re}

\begin{figure*}[h]
\begin{center}
\includegraphics[width=1.0\textwidth]{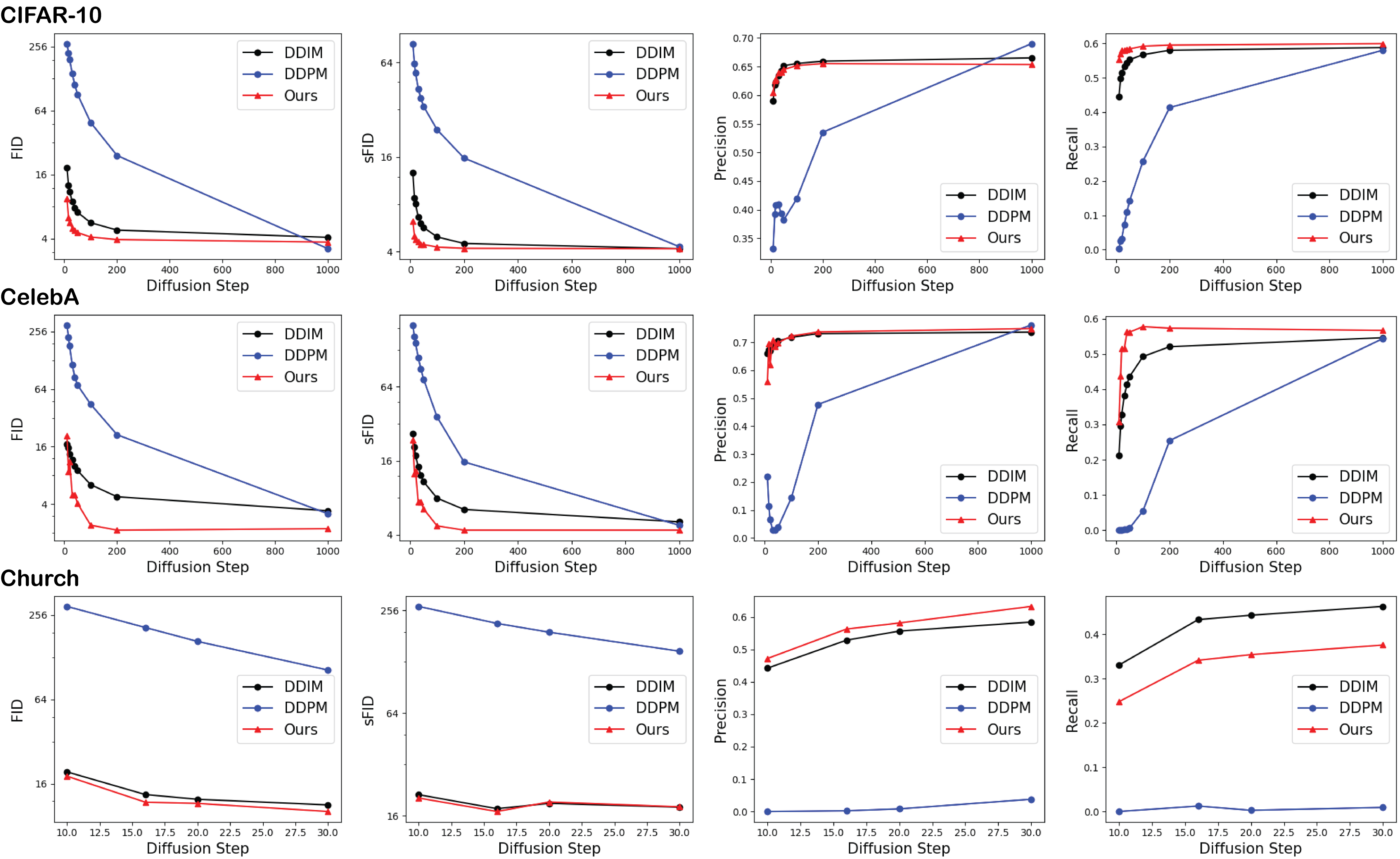}
\caption{Comparing the performance of DDPM, DDIM and our method in FID$\downarrow$, sFID$\downarrow$, Precision$\uparrow$, and Recall$\uparrow$ evaluation metrics in three different scale datasets, CIFAR-10 (32$\times$32), CelebA (64$\times$64) and LUSH outdoor Church (256$\times$256). \textbf{Top:} Measuring the metrics value change with the diffusion steps increase in the CIFAR-10 dataset. \textbf{Middle:} CelebA dataset. \textbf{Bottom:} LUSH outdoor church.} \label{fig:result}
\end{center}
\end{figure*}

\begin{figure}[h]
\begin{center}
\includegraphics[width=0.9\textwidth]{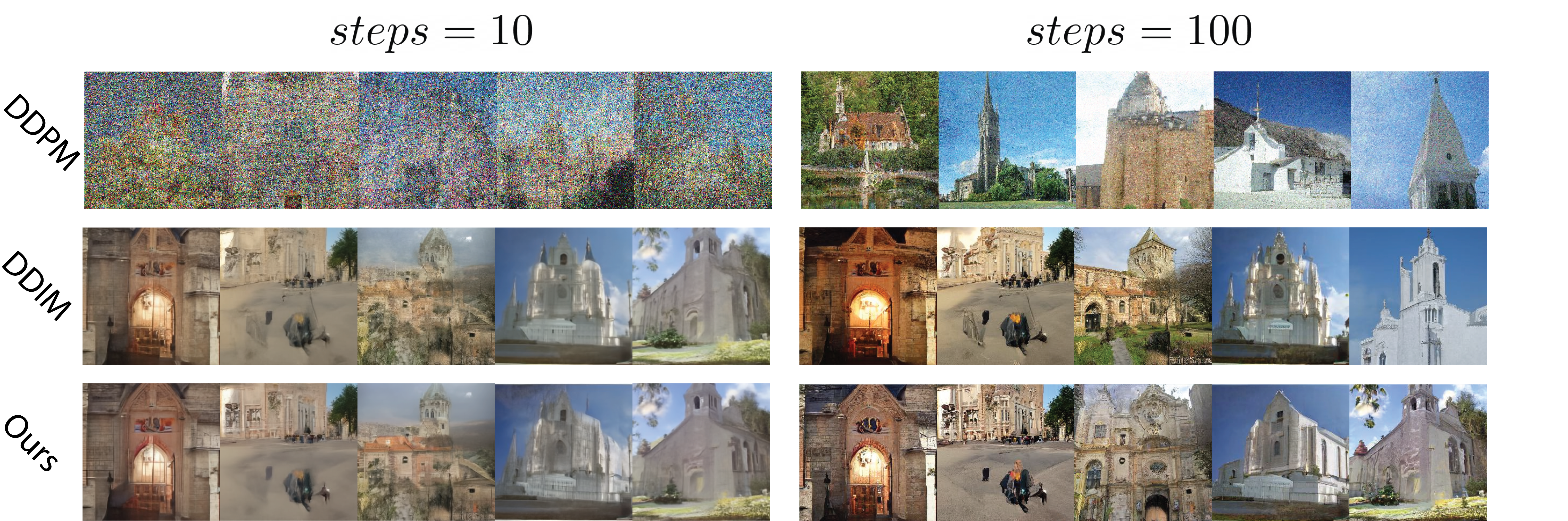}
\caption{LUSH outdoor Church (256$\times$256) samples with sampling step $10, 100$, and \textbf{Top:} is DDPM with a pre-trained model with 4432000 iterations, \textbf{Middle:} is DDIM with the same pre-trained model with DDPM, \textbf{Bottom:} is Our method \textit{without using gradient update} in the sampling process and the model's iterations are 1135000.}\label{fig:church}
\end{center}
\end{figure}

In \figureref{fig:result} and \tableref{table:seed}, we conducted a comparison of sample quality from DDPM~\cite{ho2020denoising}, DDIM~\cite{song2020denoising}, and our model trained on CIFAR10, CelebA, and Church datasets. The samples were evaluated based on FID, sFID, precision, and recall, with variation in the number of time steps used for sample generation. In this figure, for the FID score and sFID score, we can find that our model outperforms DDPM and DDIM on all these three datasets, and the improvement is especially noticeable when the steps are between 50 and 200. However, when the number of steps reaches 1000, in the CIFAR-10 dataset, the FID value in DDPM is better than our model, since DDPM tends to get better performance than DDIM when the maximum number of steps for training is the same as the number of steps for inference. For the score of precision and recall, our model can outperform the other two models in both CIFAR-10 and CelebA datasets, which means that the accuracy and diversity of the results predicted by our model are better. For the Church dataset, we observe that its results are similar to those of DDIM. Combined with the experiments in \figureref{fig:church}, we use these results to show that our model requires less training time to obtain results similar to those of DDIM.


\begin{figure}[h]
\begin{center}
\includegraphics[width=1.0\textwidth]{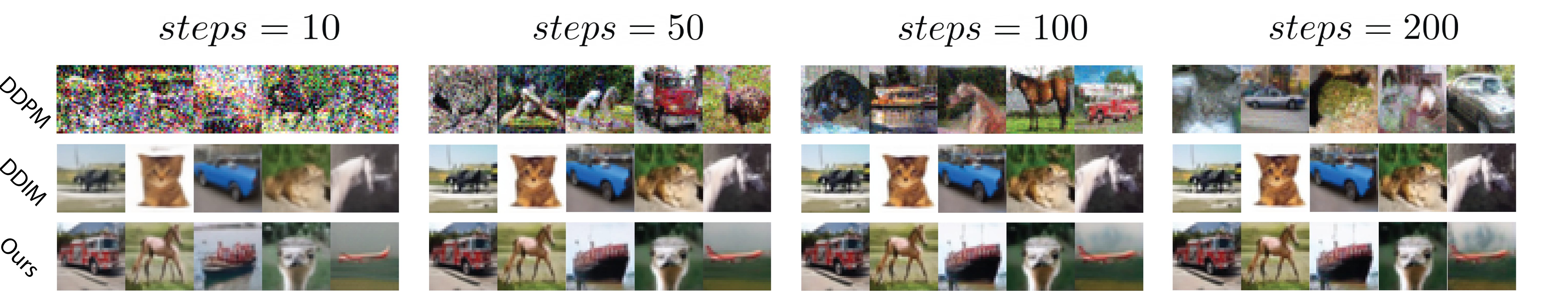}
\caption{CIFAR-10 (32$\times$32) samples with sampling steps $10, 50, 100, 200$.}\label{fig:cifar}
\end{center}
\end{figure}

\begin{figure}[h]
\begin{center}
\includegraphics[width=1.0\textwidth]{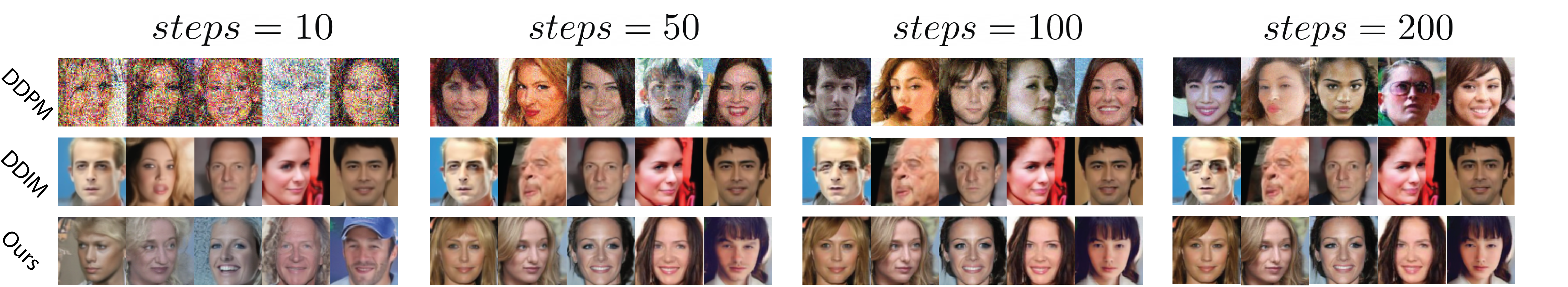}
\caption{CelebA (64$\times$64) samples with sampling steps $10, 50, 100, 200$.}\label{fig:celeba}
\end{center}
\end{figure}

In \figureref{fig:cifar,fig:celeba,fig:church},  we present the images of CIFAR-10, CelebA, Church sampled in DDPM~\cite{ho2020denoising} model, DDIM~\cite{song2020denoising} model, and our method using the same number of sampling steps. The results show that, if the sampling steps are 10, the quality of DDPM's sampling is notably inadequate. DDIM's sampling quality is considerably better than DDPM, although the image appears slightly blurred. However, our model can achieve clearer outcomes than those of DDIM. Furthermore, we have made an additional noteworthy observation through experiments. As the dimensions of training images increase, the complexity of the training process also escalates. Models trained entirely on noise, like DDIM~\cite{song2020denoising}, and DDPM~\cite{ho2020denoising}, require longer training times, e.g., at least 4,432,000 iterations for good performance on the LUSH outdoor Church dataset. However, in our training phase, we improve the performance of the model by simultaneously learning the underlying distribution of the images directly, which can be observed in the \figureref{fig:church} that our model only needs 1,135,000 iterations to obtain similar performance to DDIM and DDPM.

\begin{figure}[h]
\begin{center}
\includegraphics[width=1.0\textwidth]{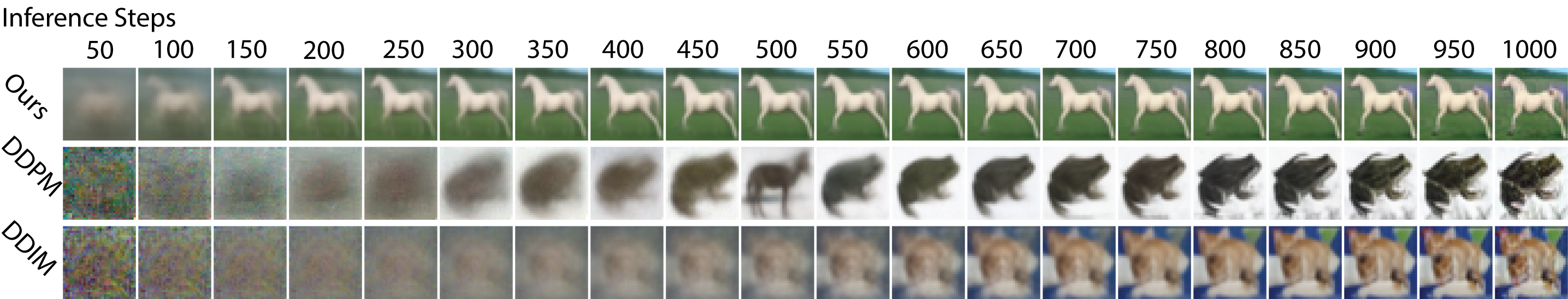}
\caption{Example of a CIFAR-10 image estimation ($\hat{x_0}$ ) over 1000 sampling steps from three models (Ours, DDPM, DDIM). These three results are derived from the same noise input, which leads to different results due to different model parameters and sampling methods.}\label{fig:prog}
\end{center}
\end{figure}


In addition, \figureref{fig:prog} shows the advantage of our model over traditional models in that it can be converted from noise to coarse realistic images much faster. From this figure, the image generated from our model can be clearly seen as the object of a "horse" by taking roughly 150 steps, but the same process in DDIM or DDPM has to spend roughly 400 to 500 steps. This means that during sampling, our model can convert from pure noise to normal images 3 times faster than DDPM and DDIM models.

\begin{figure}[htp]
\begin{center}
\includegraphics[width=0.9\textwidth]{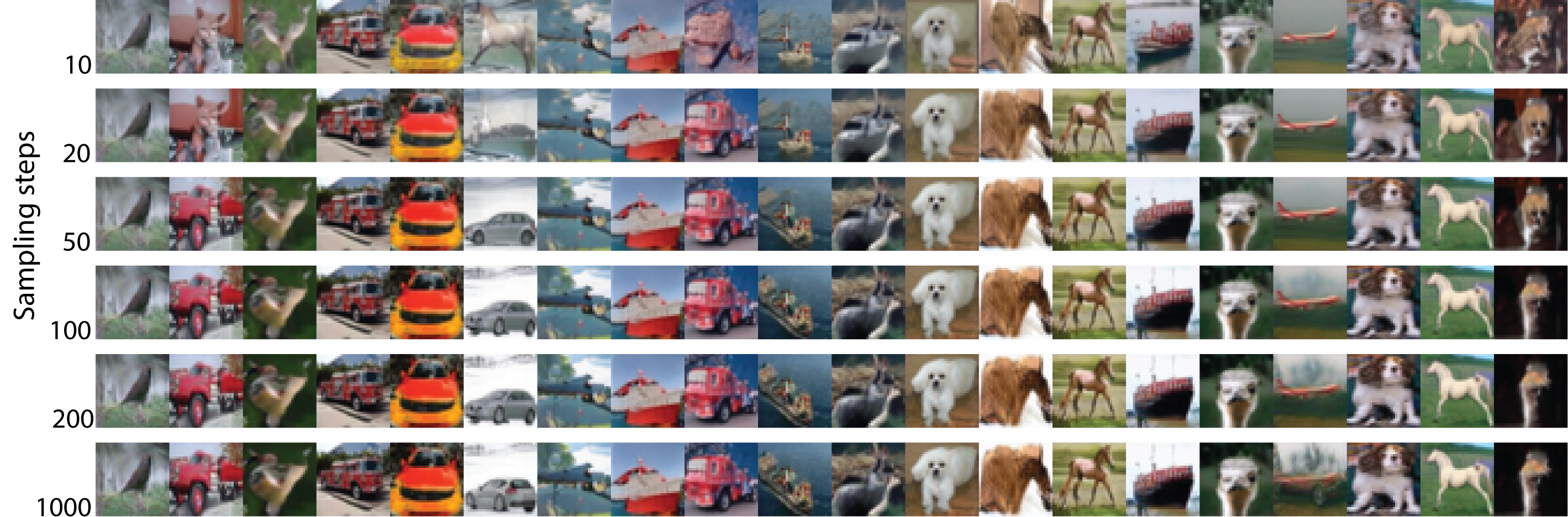}
\caption{Sampling from our model in different sampling step in CIFAR-10 dataset.}\label{fig:models}
\end{center}
\end{figure}

\begin{table}[]
\centering
\caption{To verify the stability of our model, we randomly selected three random seeds and obtained an interval of the FID scores.} \label{table:seed}
\begin{tabular}{l|ccccc}
\hline
\multicolumn{1}{c|}{Steps} & 10    & 20    & 50   & 100 & 1000 \\ \hline
CIFAR-10                   & $9.50\pm0.10$  & $5.62\pm0.04$  & $4.57\pm0.02$ & $4.17\pm0.05$ & $3.70\pm0.03$\\
CelebA                     & $22.70\pm0.12$ & $11.00\pm0.10$ & $4.05\pm0.05$ & $2.39\pm0.03$ & $2.20\pm0.02$\\ \hline
\end{tabular}
\end{table}
Moreover, in \figureref{fig:models}, we presented the results obtained at different sampling steps from our model.
\subsection{Ablation Experiments}
\label{exp:al}
We show the results of ablation experiments for our proposed three contributions, first is a novel noise schedule, the second is a new parameterization method, which is described in Section~\ref{meth:1}, and the last is a method to estimate both noise and image simultaneously mentioned in Section~\ref{meth:2}. Our ablation experiments are based on modifying the DDIM reference code, and we treat the original DDIM model as a baseline. To compare the performance of each contribution, we use FID as the evaluation metric. 

\begin{table}[]
\caption{DDIM+$\beta^*$ denotes the first contribution, which proposes a novel noise schedule. DDIM+$\sin()$ represents the new parameterization method with $\cos(\cdot)\mathbf{x}_0 + \sin(\cdot)\mathbf{\epsilon}$. Additionally, the DDIM+$\hat{\mathbf{x}}_0, \hat{\mathbf{\epsilon}}$ indicates the method of estimating noise and image both. Ours represents our final model, which incorporates the contributions ($\beta^*$ and $\hat{\mathbf{x}}_0, \hat{\mathbf{\epsilon}}$) mentioned earlier and utilizes gradient information for image sampling.} \label{ablation}
\begin{tabular}{l|ccccc|ccccc|}
                   & \multicolumn{5}{c|}{CIFAR-10}                                                 & \multicolumn{5}{c|}{CelebA}                                                     \\
Steps              & 10            & 16            & 20            & 50            & 100           & 10             & 16            & 20             & 50            & 100           \\ \hline
\footnotesize{DDIM}               & 18.67         & 12.67         & 11.06         & 7.08          & 5.64          & 16.92          & 15.68         & 13.38          & 9.00          & 6.40          \\
\footnotesize{DDIM+$\beta^*$}    & 15.04         & 10.77         & 9.28          & 6.00          & 4.92          & 23.03          & 15.23         & 13.06          & 5.92          & 4.24          \\
\footnotesize{DDIM+$\sin()$} & 15.97         & 11.90         & 9.79          & 6.22          & 5.10          & 24.71          & 17.18         & 11.87          & 5.25          & 4.22          \\
\footnotesize{DDIM+$\hat{\mathbf{x}}_0, \hat{\mathbf{\epsilon}}$}       & 15.03         & 10.05         & 8.84          & 6.25          & 5.36          & \textbf{14.94} & 13.01         & 11.56          & 8.08          & 6.11          \\ \hline
\begin{tabular}[c]{@{}l@{}}\footnotesize{DDIM+$\sin()$}\\           \footnotesize{\qquad\;\;\;+$\hat{\mathbf{x}}_0, \hat{\mathbf{\epsilon}}$}\end{tabular} &  13.17 &  9.77& 8.29 &  5.50& 4.50 & 22.56 & 13.89 & 12.80 & 5.05 & 3.09 \\ 
\begin{tabular}[c]{@{}l@{}}\footnotesize{DDIM+$\beta^*$}\\           \footnotesize{\qquad\;\;\;+$\hat{\mathbf{x}}_0, \hat{\mathbf{\epsilon}}$}\end{tabular} &  13.69 &  10.19 & 8.73 &  5.80 & 4.73 & 23.92 & 12.58 & 12.67 & 4.72 & 2.83 \\ \hline

Ours               & \textbf{9.50} & \textbf{6.32} & \textbf{5.62} & \textbf{4.57} & \textbf{4.17} & 22.70          & \textbf{8.80} & \textbf{11.00} & \textbf{4.05} & \textbf{2.39}
\end{tabular}
\end{table}

In \tableref{ablation}, we present the ablation experiments conducted to demonstrate the three contributions of our proposed approach. The baseline model used for comparison is DDIM~\cite{song2020denoising}. From the result, we found that each contribution results in a lower FID~\cite{heusel2017gans} value compared to the baseline model, which means the generated images from our model are better than the base model in the CIFAR-10 and CelebA datasets. Meanwhile, we obtained further improvements to the model by combining the three contributions discussed above. From the table, it can be observed that compared to models that only utilize individual contributions, the performance is further enhanced by combining $\sin()$ contribution with $\hat{\mathbf{x}}_0, \hat{\mathbf{\epsilon}}$ contribution, as well as by combining $\beta^*$ contribution with $\hat{\mathbf{x}}_0, \hat{\mathbf{\epsilon}}$ contribution. However, since both $\beta^*$ and $\sin()$ contributions aim at optimizing the efficiency and performance of training and sampling for diffusion models by improving the noise schedule, we do not combine them as a comparison model. Furthermore, in our final model, the use of $\sin()$ makes it more convenient to transform the parameterization equation into an ODE, and it is beneficial to apply the gradient approach as the sampling approach. Thus, our final model consists of $\sin()$, $\hat{\mathbf{x}}_0, \hat{\mathbf{\epsilon}}$ and a gradient update-based sampling process, which achieves the best result compared to the aforementioned models.

Meanwhile, by comparing the results of DDIM+$\hat{\mathbf{x}}_0, \hat{\mathbf{\epsilon}}$ with those in DDIM+$\beta^*$ and DDIM+$\sin()$, we find this result tends to be better when the sampling steps are less than 20, but when the steps are greater than 20, the impact of DDIM+$\hat{\mathbf{x}}_0, \hat{\mathbf{\epsilon}}$ is lower than the impact of the other two contributions. Based on our analysis, it is evident that using the conventional linear noise schedule in DDIM+$\hat{\mathbf{x}}_0, \hat{\mathbf{\epsilon}}$ model results in the noise contribution outweighing the contribution from the image in the loss function. This is primarily due to the concentration of training on the noisy regions within the linear noise schedule. This is also demonstrated by the results of our DDIM+$\beta^*$+$\hat{\mathbf{x}}_0, \hat{\mathbf{\epsilon}}$ model, and DDIM+$\sin()$+$\hat{\mathbf{x}}_0, \hat{\mathbf{\epsilon}}$ model. When we use the new noise schedule in the DDIM+$\hat{\mathbf{x}}_0, \hat{\mathbf{\epsilon}}$ model, this solves the previous problem that performance deteriorates when steps are increased, through balancing the contribution of noise and image components.

\section{Conclusion}

In this work, our paper introduces two key contributions to the current diffusion methods, resulting in significant improvements in the quality and diversity of generated results. Through applying these two contributions, our model achieves faster generation, with the ability to converge on high-quality images more quickly, and higher quality of the generated images than DDPM, DDIM, and Cold Diffusion models. Our model increases the controllability of the generation process by learning both noise and image information and using gradient information in the inverse diffusion process.

\acks{This research was supported by The University of Melbourne’s Research Computing Services and the Petascale Campus Initiative.}

\bibliography{acml23}

\end{document}